# An Adaptive Simulated Annealing-Based Machine Learning Approach for Developing an E-Triage Tool for Hospital Emergency Operations

Abdulaziz Ahmed[a], Mohammed Al-Maamari[b], Mohammad Firouz[c], Dursun Delen[d, e*]

**Abstract**

Patient triage at emergency departments (EDs) is necessary to prioritize care for patients with critical and time-sensitive conditions. Different tools are used for patient triage and one of the most common ones is the emergency severity index (ESI), which has a scale of five levels, where level 1 is the most urgent and level 5 is the least urgent. This paper proposes a framework for utilizing machine learning to develop an e-triage tool that can be used at EDs. A large retrospective dataset of ED patient visits is obtained from the electronic health record of a healthcare provider in the Midwest of the US for three years. However, the main challenge of using machine learning algorithms is that most of them have many parameters and without optimizing these parameters, developing a high-performance model is not possible. This paper proposes an approach to optimize the hyperparameters of machine learning. The metaheuristic optimization algorithms simulated annealing (SA) and adaptive simulated annealing (ASA) are proposed to optimize the parameters of extreme gradient boosting (XGB) and categorical boosting (CaB). The newly proposed algorithms are SA-XGB, ASA-XGB, SA-CaB, ASA-CaB. Grid search (GS), which is a traditional approach used for machine learning fine-tunning is also used to fine-tune the parameters of XGB and CaB, which are named GS-XGB and GS-CaB. The six algorithms are trained and tested using eight data groups obtained from the feature selection phase. The results show ASA-CaB outperformed all the proposed algorithms with accuracy, precision, recall, and f1 of 83.3%, 83.2%, 83.3%, 83.2%, respectively.

## 1. Introduction

* Corresponding author

Phone: +1 (918) 594-8283

Email: dursun.delen@okstate.edu

Emergency department (ED) visits have increased significantly over the last two decades. In 2000, there were around 108 million emergency visits in the United States [1]. Over the past twenty years, the number of emergency visits has increased to 143 million [2]. Prioritizing ED patients is a critical process due to the limited resources and the large volume of ED visits. One of the most common algorithms to prioritize ED patients is the Emergency Severity Index (ESI). It is used by approximately 80% of the hospitals in the United States [3]. It has five levels (1-5) to represent the urgency of a patient ED visit, in which level-1 is the most urgent and level-5 is the least urgent. Inaccurate assignment of ESI level leads to under-triage or over-triage of ED patients, which results in either over utilizing ED resources or putting a patient's life at risk [4]. For example, if a patient should be assigned level-2 and level-3 is assigned instead (e.g., under-triage), the patient's life could be in danger as level-3 implies that the patient would wait roughly 30 minutes [5]. In 2008, at the Vista Medical Center Emergency Room in Lake County, Illinois, a nurse under-triaged a patient who came to the ED complaining of shortness of breath, nausea, and serious chest pain. The nurse sent the patient to the ED waiting room, after waiting for two hours, the patient was found dead [6].On the other hand, over-triage wastes scarce resources that could be used to treat other patients who are in more need of these resources the most [7]. Therefore, there is a critical need to develop an approach for an efficient and accurate triage process, which would save lives, improve resource utilization at EDs.

Over the past ten years, machine learning algorithms have been developed well and used in a wide variety of health care fields, including preventive medicine [8], cancer diagnosis [9], and organ transplant [10] 2021). In the case of ED triage, developing fast and accurate machine learning models can improve triage accuracy. Due to the ED overcrowding problem, nurses are overwhelmed and stressed, especially in the era of pandemics (e.g., COVID-19) [11]. As a result, it is difficult for nurses to always adhere to the triage guidelines and make accurate triage decisions. When nurses are not able to follow ESI guidelines, an under-triage is more likely to occur [12]. To reduce stress on triage nurses and avoid inaccurate triage, a machine learning model can be utilized to develop an e-triage tool that can help nurses to make accurate triage decisions and consequently minimize the rates of under-triage and over-triage [13]. Machine learning approaches have been proven in the literature to be effective in developing accurate models due to their ability to identify non-linear relationships between different variables. Machine learning also can handle numerous features and work with high dimensional and complex data [13,14]. However, one of the main challenges of utilizing machine learning algorithms is that most of them have many parameters and in many cases, there are infinite possible values for these parameters. For example, the learning rate in Extreme Gradient Boosting (XGB) can have any value between 0 and 1, while the number of estimators can have any value between 1 and $\infty$. Without finding the optimal values of the parameters of machine learning algorithms, obtaining a high accuracy model becomes difficult.

The purpose of this study is to develop an e-triage tool based on two main algorithms: Extreme Gradient Boosting (XGB) and Categorical Boosting (CaB). To improve the performance of the two algorithms, the current study proposes a framework for integrating metaheuristic optimization algorithms, Simulated Annealing (SA) and Adaptive Simulated Annealing (ASA) with XGB and CaB machine learning methods. The purpose of SA and ASA algorithms is to optimize the parameters of XGB and CaB prediction methods. The newly proposed algorithms, SA-XGB, SA-CaB, ASA-XGB, and ASA-CaB. The SA-XGB, and SA-CaB, are based on SA while ASA-XGB and ASA-CaB are based on ASA. Although SA was used for optimizing the number of nodes in hidden layers for deep neural networks [17], to the best of our knowledge, this is the first study that proposes the use of ASA for optimizing decision tree ensembles (i.e., XGB and CaB). The proposed framework can be generalized and used for any machine learning algorithm. For comparison, grid search (GS) is also utilized to fine-tune the parameters of XGB and CaB, which are named GS-XGB and GS-CaB, respectively. A large dataset is collected from different ED locations of a large hospital located in the Midwest of the United States. The modeling involves three phases: data preprocessing, feature selection, and model development. Decision tree (DT), random forest (RF), and Chi-square (Chi-sq) are utilized for feature selection and implemented using three search algorithms: select K best (SKB), recursive feature elimination (RFE), and select from the model (SFM). In the model development, the data groups obtained from the feature selection phase are used to train and test the proposed algorithms (e.g., SA-XGB, SA-CaB, ASA-XGB, and ASA-CaB, GS-XGB, and GS-CaB). The accuracy, sensitivity, specificity, and f1 score are used as the performance measures to evaluate the proposed models. The most-optimized model can be used as an e-triage tool that can be used by nurses to prioritize the care of ED patients to reduce the risk of over and under triage, and consequently, improve patient flow. This study contributes to the theory and practice of healthcare analytics, the application of which creates value to healthcare operations. This study also contributes to the area of integrating machine learning and optimization by highlighting how machine learning performance can be boosted based on optimization concepts.

The remainder of the manuscript is organized as follows: Section 2 provides an overview of the studies that utilized machine learning for predicting ESI and the studies that integrated machine learning and metaheuristic optimization. Section 3 presents the proposed methodologies, while the results are presented in Section 4. The findings and managerial impacts of the proposed models are discussed in Section 5. The conclusion and future work are presented in Section 6.

## 2. Literature Review

In this section, we first review the literature on using machine learning for predicting of ESI, and then we proceed with the existing works on using optimization methods for improving machine learning techniques.

### 2.1 Predicting ESI using machine learning

ESI is a triage tool that is used in several countries including the U.S., Canada, and Australia [18]. Wuerz at al. [19] developed the first version of the ESI tool and then kept refining and updating it. In 2019, the Emergency Nurses Association (ENA) acquired it [20]. The most updated version of the ESI tool is version 4 and it has 5 levels, in which level 1 is the most urgent and level 5 is the least urgent. The five ESI can be merged into three levels. Levels 1 and 2 can be merged into one class (most urgent), while levels 4 and 5 can be merged into one class (less urgent). Level 3 can be standalone (urgent) [7]. The rationale behind this merging is based on clinical reasons. levels 1 and 2 are time-sensitive as a patient with those levels must be treated immediately. A Level 3 patient often waits hours, while levels 4 and 5 are categorized in separate areas. Therefore, under-triaging a patient from level 4 to level 5 would have a much smaller effect than under-triaging a patient from level 3 to level 4 [7]. In this study, the ESI levels (1, 2) were merged into one class (most urgent), levels (3, 4) into (urgent), and level-5 (less urgent). From a modeling perspective, the main reason for the merging is to mitigate the severe data imbalance among the ESI levels in the dataset we used for training the proposed models. In this study, three levels are considered levels 1 and 2 (most urgent), level 3 (urgent), and level 4 and 5 (less urgent).

Various studies have used machine learning to predict ESI levels of ED patients [21–23]. Other studies used machine learning to develop binary ESI by merging levels 1 and 2 into one class, which indicates (critical), and levels (3, 4, 5) into another class, which indicates (less urgent or uncritical) [24–26]. In this study, three levels are considered levels 1, 2 (most urgent), level 3 (urgent), and level 4, 5 (less urgent). Regarding the predictors of ESI, some studies used triage information of ED patients (e.g., triage vital signs, chief complaints, demographics) [21–23,25,26]. Other studies included patients' medical history in addition to triage information including kidney disease, diabetes, and heart disease [24,27]. The number of patients included in previous studies varied from approximately 950 patients to about 800,000 patients. Klug at al. [27] used the largest dataset available to predict ESI, which included 799,522 ED visits. Chonde at al. [21] used only 947 visits, while the size of the datasets used by [22,23,25,26] are 550k, 486k, 445k, 172k, 147k, and 135k ED visits, respectively. The dataset size used for this study includes about 450K ED visits.

When it comes to the methodologies that are used for predicting ESI, different machine learning algorithms were utilized. Such algorithms are XGB, Gradient Boosting [27], and Random Forest [24]. Other studies exploited more than one model for prediction. Chonde et al. [21] used Ordinal Logistic Regression (OLR), Naïve Bayesian Networks (NBNs), and Sekandari and Saleh [22] used Artificial Neural Networks (NNs), RF, Gaussian Naïve Bayes, and Gradient Boosting to predict ESI. When it comes to hyperparameter optimization, Joseph at al. [26] utilized random search (RS) to optimize the parameter of their DNN model, while Sekandari and Saleh [22] used GS for optimizing the parameters RF. In this paper, seven feature selection methods are utilized. Also, six algorithms are proposed to build a model to predict ESI, which are SA-XGB, ASA-XGB, SA-CaB, ASA-CaB, GS-XGB, and GS-CaB. Although SA has been used to optimize other algorithms (e.g., DNN) (See Table 1), up to our knowledge, this is the first study that proposes the use of ASA to optimize machine learning and more specifically, XGB and CaB.

## 2.2 Machine learning parameter optimization

Different methods are used for optimizing the hyperparameters of machine learning algorithms. Table 1 summarizes the approaches that have been proposed in the literature to optimize the hyperparameters of machine learning including GS, RS [28,29], Bayesian optimization (BO) [30], and metaheuristic. In GS, hyperparameter values are mapped as a grid space. Then, a model is trained and evaluated based on every single value in the grid. In other words, all possible combinations of different hyperparameters are tested. The disadvantage of GS is that as the number of parameters increases, the number of times a model gets trained and tested increases exponentially, which makes it time-consuming. In GS, obtaining the optimal hyperparameters for a machine learning model is not guaranteed [31]. In RS, the search space of a model's parameters is pre-defined by upper and lower bounds of hyperparameters values. Then, the different values for the hyperparameters are chosen randomly and then evaluated. However, the RS method has high variance and does not guarantee optimality [32]. BO is also used for optimizing machine learning parameters [30,33]. The disadvantage of BO is that it quickly degrades as the number of parameters increases.

In addition to GS, RS, and BO, metaheuristic optimization approaches such as genetic algorithm (GA) and simulated annealing (SA) were utilized to optimize the hyperparameters of different machine learning algorithms. The advantage of metaheuristic optimization algorithms is that they can solve non-convex, discrete, and non-smooth optimization problems. Some studies used population-based algorithms, while others used single solution-based algorithms. Particle swarm optimization (PSO) and GA, which are population-based algorithms, are the most used methods. PSO and GA were utilized to find the optimal value of the regularization parameter (C) of SVM [34,35] and the number of hidden neurons and learning in artificial neural networks (ANN) [14]. The hyperparameters of XGB were optimized using GA [36]. Other population-based algorithms including Artificial Bee Colony (ABC), Ant Lion Optimization (ALO), and Bat Algorithm (BA) were used to optimize a few hyperparameters of a convolutional neural network (CNN) [37]. Also, moth flame optimization (MFO), gray wolf optimization (GWO), and whale optimization algorithm (WOA) are used to optimize the parameters of SVM [38]. Differential Flower Pollination (DFP) was used to optimize an SVM model [39]. The main disadvantage of GA and PSO is that they have expensive computational costs since they are population-based metaheuristics [40]. Single solution-based algorithms were utilized for hyperparameter optimization for machine learning. The SA algorithm was to determine the optimal number of hidden layers in DNN, which was the [41]. Tabu search (TS) was utilized to find the optimal hyperparameters of Adaboost (ADAB) [42]. TS was also used to optimize the hyperparameter of XGB, ADAB, and neural networks (LP) [43].

In addition to GS, RS, BO, and metaheuristics frameworks, there are existing libraries that can be used for hyperparameter optimization. Such libraries are Spearmint, BayesOpt, Hyperopt, SMAC, and MOE. These frameworks are based on RS, BO, or a combination of both. More information about these frameworks can be obtained from the work presented by [44]. Our study proposes a new metaheuristics optimization algorithm for optimizing machine learning parameters. This work's contributions include:

- We investigate the parameter optimization problem which is a major issue machine learning algorithms.

- The problem of hyperparameter fine-tuning is modeled as an optimization problem and a framework is proposed to show how SA and ASA can be used to optimize XGB and CaB.

- Although SA was used for optimizing the parameters of DNN, up to our knowledge, this study is the first study that proposes ASA for optimizing the hyperparameters of machine learning (XGB and CaB).

- The proposed algorithms are implemented on a multi-classification problem, which is predicting the ESI levels for emergency patients. The purpose of our study is to develop an accurate model with a minimum number of features to predict three ESI levels.

- An extensive number of experiments are conducted to show the robustness of the proposed algorithms.

**Table 1:** Summary of studies that proposed methods for optimizing hyperparameters of machine learning.

| Author(s) | Machine learning Algorithm(s) | GS | RS | BO | TS | GA | PSO | ALO | BA | ABC | DFP | MFO | GWO | WOA | SA | ASA |
|---|---|---|---|---|---|---|---|---|---|---|---|---|---|---|---|---|
| Bergstra et al. [28] | DBNs | | * | | | | | | | | | | | | | |
| Bergstra & [31] | DBNs | | * | | | | | | | | | | | | | |
| Guo et al. [30] | XGB | | | * | | | | | | | | | | | | |

| | | | | | | | | | | | | | | | |
|---|---|---|---|---|---|---|---|---|---|---|---|---|---|---|---|
| Badrouchi et al. [10] | LR, KNN, XGB, MLP | * | | | | | | | | | | | | | |
| Chou et al. [35] | SVMs | | | | | * | | | | | | | | | |
| Pham & Triantaphyllou [34] | SVM, ANN, DT | | | | | * | | | | | | | | | |
| Sarkar et al. [14] | SVM, ANN | | | | | * | * | | | | | | | | |
| Chen et al. [36] | XGB | | | | | * | | | | | | | | | |
| Tsai et al. [17] | CNN | | | | | | | | | | | | | * | |
| Bereta [42] | Adaboost | | | | * | | | | | | | | | | |
| Gaspar et al. (2021) [37] | CNN | | | | | | * | * | * | * | * | | | | |
| Bibaeva [45] | CNN | | | | | * | | | | | | | | * | |
| Hoang & Tran [39] | SVM | | | | | | | | | | | | * | | |
| Guo et al. [46] | DNN | | | | * | * | | | | | | | | | |
| Snoek et al. [33] | NN | | | * | | | | | | | | | | | |
| Zhou et al. [38] | SVM | | | | | | | | | | | | * | * | * |
| Ahmed at al. [43] | XGB, ADAB, MLP | * | | | * | | | | | | | | | | |
| **Our study** | **XGB, CaB** | * | | | | | | | | | | | | * | * |

# 3 Research Methodology

The methodologies developed in this study including the proposed framework, data preprocessing, feature selection, and model development are presented in this section.

## 3.1 The proposed framework

Figure 1 illustrates the proposed framework. In phase I, the data is preprocessed by handling missing values and removing unnecessary features, which are the ones that occur after patient triage. The categorical features such as gender and race are encoded. In phase II, feature selection is conducted. Four feature selection methods are used, which are regularized multinomial logistic regression (Lasso), RF, DT, and Chi-sq. These algorithms are implemented using three feature search algorithms: select from the model (SFM), recursive feature elimination (RFE), and select $K$ best (SKB). SFM and RFE are fitted using Lasso, RF, and DT, while SKB implemented Chi-sq. The feature selection step results in seven data groups: (1) Lasso_SFM, (2) RF_SFM, (3) DT_SFM, (4) Chi-sq_SKB, (5) Lasso_RFE, (6) RF_RFE, and (7) DT_RFE. These seven data groups and a group with all the features, which is denoted by X_all, are utilized to build the prediction model using the newly proposed algorithms: SA-XGB, SA-CaB, ASA-XGB, ASA-CaB, GS-XGB, and GS-CaB. In summary, 48 models are developed ((7 data groups resulted from feature selection step + all features) × 6 prediction methods = 48 models). The hyperparameters in SA-XGB and SA-CaB are optimized using SA, while the parameters of ASA-XGB and ASA-CaB are optimized using ASA. The parameters of GS-XGB and GS-CaB are fine-tuned using GS. Oversampling is utilized to solve the data imbalance problem using the Synthetic Minority Oversampling Technique (SMOTE). During the optimization stage, accuracy is used as the main performance metric. Five performance measures are used: accuracy, precision, recall, and f1 score.

## 3.2 Data Acquisition and Preprocessing

The retrospective data for this study is obtained from a large hospital system in the Midwest of the United States. Patient emergency records from four large locations were used in this study. The records were collected between 2017 and 2019. At first, the entire dataset has about 470k patient records and about 32 predictors. These features represented all the events that took place during an ED visit, including nurse assessment, medical diagnoses, and triage score. The following criteria are considered for including/ excluding the features considered in this study:

- All the features that describe events that occur after triaging a patient are excluded. The reason is that we aim to develop a model that predicts the ESI levels of ED patients, which means that any feature that happens after a patient is triaged is irrelevant. This includes all the time-related features except arrival time, waiting time for screening, discharge and admission time, medical diagnoses, and disposition status.

- The arrival time feature is split into multiple features: month (e.g., March, April), weekday (e.g., Thursday, Friday), and hours (e.g., 1-24). The minutes and the seconds are excluded.

- The pain scale feature is removed because it has more than 90% missing values.

After implementing the above criteria, the final number of features becomes 18 including the output feature (e.g., ESI) (See Table 2). The dataset includes many missing values for different features. The numbers and percentages of missing data for each row are shown in Table 2. $k$-Nearest Neighbor (KNN) is used to impute missing values to prevent losing important information by dropping all visits that have missing values. When KNN is used for imputation, the nearest observations for the patient with missing value are found based on Euclidean distance. Then, the missing values are filled with the average of the nearest neighbors. KNN imputer is only used with numerical features. The categorical features of the dataset used for this study do not include missing values (See Table 2). Since the data includes numerical and categorical features (See Table 2), data encoding is applied to the categorical features using integer encoding. Each categorial value in a categorical feature is given an integer number. The dataset is also scaled before training the proposed algorithms. After preprocessing the data, the sample size is 478, 212 patient visits, and 18 predictors. Since the sample size is too large, a random sample of 5000 is selected from the record to develop the proposed models.

Figure 2 shows that the output feature (i.e., ESI score) is severely imbalanced as scores one and five represent a very small number of visits compared with scores 2, 3, and 4. This will affect the accuracy of the proposed model. Therefore, level 1 is combined with level 2, while level 5 is combined with level 4. Level 3 is not merged with any level. This action is justified clinically [7]. Figure 3 shows the ESI levels that are considered in this study.

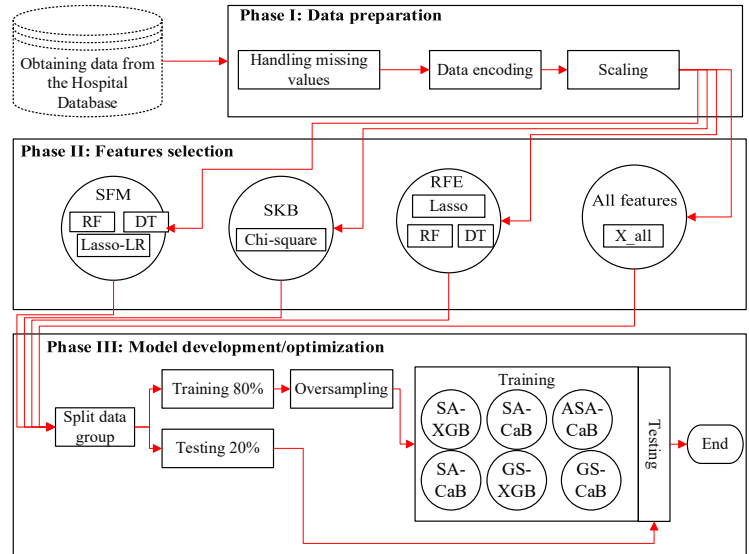

**Figure 1:** The research framework.

**Table 2:** Percentage of missing values.

| Feature | Percentage of missing values |
|---|---|
| Respiratory Rate | 27.2% |
| O2 Saturation | 26.9% |
| Body Mass Index (BMI) | 25.7% |
| Systolic Blood Pressure | 25.7% |
| Diastolic Blood Pressure | 25.7% |
| Pulse Rate | 25.7% |
| Temperature in Fahrenheit | 25.7% |
| Patient Sex | 0.0% |
| Ed Department Location ID | 0.0% |
| ED Arrival Time hour | 0.0% |
| Zipcode | 0.0% |

| Patient Ethnicity | 0.0% |
|---|---|
| Patient Smoking Status | 0.0% |
| Month of year | 0.0% |
| Day of week | 0.0% |
| Chief Complaint Topic | 0.0% |

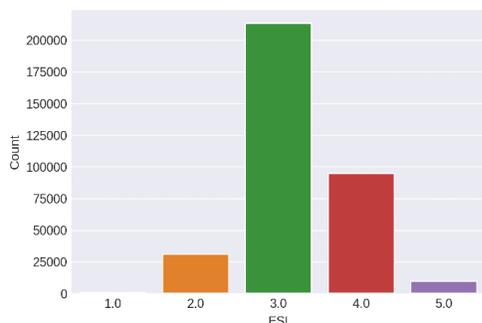

**Figure 2:** ESI score frequency for all the five ESI levels.

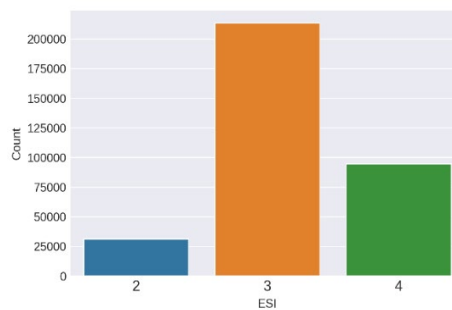

**Figure 3:** Considered ESI score frequency.

### 3.3 Modeling approaches

This section presents the methodologies used or feature selection methods, prediction algorithms as well as optimization approaches.

#### 3.3.1 Feature selection methods

Feature selection is an important step in building machine learning models. It removes irrelevant features and improves the performance of models, makes them easier to interpret, and reduces the computational time [47]. Feature selection methods can be categorized into three types: filter method, wrapper methods, and embedded methods [48]. In this paper, two methods are used, which are filter and wrapper methods. Filter methods use different scoring approaches to rank features such as Pearson's correlation and then only the top-scoring *K* features are kept. In this study, SKB is used with Chi-sq, so features are selected based on the Chi-sq score (Ma et al., 2017). The relationship between every feature with the output feature (e.g., ESI) is examined. Then, the features with the highest score are selected.

In wrapper methods, features are selected iteratively, in which a machine learning model is trained and tested based on a subset of features and then the features that result in the highest accuracy are selected. There are different wrapper methods including Backward feature elimination (BFE), Recursive feature elimination (RFE), and selection from the model (SFM) [49]. In this paper, both RFE and SFM are used. In RFE, models are created iteratively based on a different subset of features. The number of desired features is passed as a parameter. Then, during model training, a classifier is trained on the whole dataset and then each feature is weighted. Then, the features with the smallest weights are removed. The weight can be based on a model coefficient in a linear model or feature importance in models such as DT and RF. The process continues until the desired number of features is obtained [50,51]. In SFM, a threshold is defined, then the features with feature importance or coefficient less than the threshold are removed [52]. In this study, RFE and SFM are implemented with: Lasso, DT, and RF. When RF or DT are used as fitting algorithms, features importance is determined based on the Gini index. On the other hand, when Lasso is used as a fitting algorithm, feature importance is decided based on the coefficients of a model [49].

### 3.3.2 *Extreme gradient boosting (XGB)*

XGB was developed by [53]. It was developed to improve the computational speed of tree-based models. XGB is one of the gradient boosting machine (GBM) algorithms and has been used in classification [54], feature selection [36], and regression [55]. XGB is a tree-based model that combines the classification of weak learners while minimizing the loss function, which controls the classifier complexity as follows:

$$L(\Phi) = \sum_i l\left(\hat{y}_i, y_i\right) + \Omega(f_k) \tag{1}$$

Where the first term in Equation 1 represents the comparison between the actual $(y_i)$ and predicted$(\hat{y}_i)$ classes, while the second term is the regularization term and it can be written as follows:

$$\Omega(f_k) = \gamma\, T + \frac{1}{2}\, \lambda \|w\|^2 \tag{2}$$

Where T represents the number of tree leaves, $\gamma$ denotes the lowest reduction in the loss function that is necessary for splitting a node in the tree, $w$ represents the output score of leaves. XGB has more than 20 parameters and most of them can take many possible values. To improve the performance of XGB, we integrate it with SA and ASA. GS is also used for comparison. The parameters are the learning rate, number of estimators, the number of parallel trees, maximum depth, maximum delta step, and gamma. More information about XGB hyperparameters can be found in the algorithm documentation (*XGBoost Parameters Documentation*).

### 3.3.3 Categorical boosting (CaB)

The CaB is one of the gradients boosting algorithms that was developed specifically to handle categorical features. It can be utilized for classification [57], regression [58], and feature extraction [59]. The CaB is similar to other GBM algorithms such as XGB and LightGBM. CaB ensembles a variety of weak learners to create a strong one. However, unlike other GBM algorithms, CaB handles categorical features more efficiently with minimal overfitting and during the training process, which significantly reduces preprocessing time. One way to process categorical features is to convert them into numerical features using Target Statistics (TS). Different methods are used to estimate TS and one of them is called greedy TS, in which a value in a categorical feature is replaced by the mean of all the targets for a training sample, which can be calculated using Equation 3 as follows:

$$\hat{x}_k^i = \frac{\sum_{j=1}^n I_{\{x_j^i = x_k^i\}} \cdot y_j}{\sum_{j=1}^n I_{\{x_j^i = x_k^i\}}} \tag{3}$$

Where $I$ denotes the indicator function, $x_j^i$ represents the ith subtype predictor in the kth training sample. The disadvantage of the greedy TS is that it leads to target leakage and overfitting [60]. On the other hand, CaB embraces the ordering principle when calculating the TS to process categorical features. Ordered TS uses artificial "time" where it takes a random permutation $\sigma$ of training examples, then it uses all the available samples "history" to calculate its TS as follows:

$$\hat{x}_k^i = \frac{\sum_{j=1}^n I_{\{x_j^i = x_k^i\}} \cdot y_j + \beta p}{\sum_{j=1}^n I_{\{x_j^i = x_k^i\}} + \beta} \tag{4}$$

Where, $p$ a prior value to smooth the noise for low-frequency categories and $\beta$ is the weight of a prior value and it is between (0, 1). The CaB has many hyperparameters including learning rate, random strength, bagging temperature, tree depth, and L2 regularization. To optimize these parameters, we utilize SA, ASA, and GS.

## 3.4 Integrating SA and ASA with XGB and CaB

Different machine learning methods have different hyperparameters and some of them can have infinite possible values. For example, the number of trees in a CaB algorithm takes a value $(0, \infty)$, so there are many possible values to choose from. To solve such a problem, we model it as an optimization problem with an objective function and constraints. The accuracy of a model represents the objective function $f(\mathbf{x})$, where $\mathbf{x}$ represent the parameters that need to be optimized and x = $x_1, x_i, \ldots, x_n)$, $N = n_1, n_2, \ldots, n_n$ is the number of parameters (Equation 5). The limits of parameters can be represented by constraints (Equation 6), where $\psi_i^{upper}$ and $\psi_i^{lower}$ are the upper and lower bounds for $x_i$ parameter, respectively. We can write the optimization model as follows:

$$Max\ f(\mathbf{x}) = f\ (x_1, x_i, \ldots, x_n) \qquad (5)$$

Subject to: $\quad \psi_i^{lower} \leq x_i \leq \psi_i^{upper} \quad i = 1, 2, \ldots N \qquad (6)$

The goal is to solve the optimization problem and find the optimal values of $x_1, x_i, \ldots, x_n$ that maximize the objective function. The value $x_i$ can be float, integer, or binary. For example, in CaB, the number of estimators must integer and can be $(0, \infty)$, while the regularization parameters alpha is float and takes any value between [0, 1]. These constraints must not be violated while selecting the values of the parameters of CaB. In this paper, the SA and ASA are used to find the optimal values of the hyperparameters of both XGB and CaB.

### 3.4.1    Traditional simulated annealing

SA is a common metaheuristic algorithm that is used to solve combinatorial optimization problems. It was developed by [61]. It is inspired by the annealing procedure of metalworking. The annealing process is a heat treatment method, in which a metal is subjected to physical and sometimes chemical changes in its properties. In the annealing process, the metal is heated to above its recrystallization temperature, maintained for a specific period, and then, cooled to its solid state. The optimal arrangement of metal particles is shaped during the annealing process, which depends on the cooling rates. SA implements iterative movements based on a temperature parameter, which is similar to the annealing process in metal. Algorithm 1 shows how SA can be used to optimize the parameters of machine learning. SA starts with an initial solution $S^*$ (e.g., parameters' values). The initial solution becomes the current solution ($S_{cur}$). Then, a new solution ($S_{new}$) is generated by imposing a small change in one or a combination of variables' current solution ($S_{cur}$). The accuracy of the newly generated solution $f(S_{new})$ is compared with the accuracy of the current solution ($f(S_{cur})$). If the accuracy of the new solution is better than the accuracy of the initial solution ($f(S_{new}) > f(S_{cur})$), the new solution becomes the current solution ($S_{cur} = S_{new}$). However, if the new solution does not result in better accuracy, then the probability number ($c$) is calculated (Equation 7), where $T_i$ is the current annealing temperature. After that, $c$ is compared with a random number generated based on a uniform distribution between (0, 1). If $c$ is greater or equal to the randomly generated number, then, the new solution replaces the current solution ($S_{cur} = S_{new}$), even though it is a worse solution, which guarantees the diversity of the parameters and prevents the search from trapping in a local-maxima. Otherwise, the current solution is kept, and a new solution is generated. Traditional SA starts with a large temperature and decreases by a constant value ($\alpha$) (Equation 8).

$$c = \frac{1}{e^{\frac{f(S_{new}) - f(S_{curr})}{T_i}}} \tag{7}$$

$$T_i = T_{i-1} \cdot \alpha \tag{8}$$

| **Algorithm 1:** Simulated Annealing |
|---|
| 1    Set the initial parameters of SA |
| 2    Generate an initial solution $S^*$ |
| 3    Set $S^*$ as the current solution ($S_{cur}$) |
| 4    **While** the stopping criterion is not met |
| 5        Generate neighborhood solution ($S_{new}$) |
| 6        Evaluate $S_{new}$ |
| 7    **if** $f(S_{new}) > f(S_{cur})$ |
| 8        $S_{cur} = S_{new}$ |
| 9    **elif** $rand\,(0,1) <= c$ |
| 10        $S_{cur} = S_{new}$ |
| 11    else |
| 12        Keep $S_{cur}$ and Go to line 4 |
| 13    Update annealing temperature ($T_i = T_{i-1} \cdot \alpha$) |
| 14    End |

### 3.4.2 *Adaptive simulated annealing*

In this section, a modified version of SA is proposed to optimize machine learning and it is called adaptive SA (See Algorithm 2). The main difference between SA and ASA comes from the mechanism of changing the annealing temperature. In traditional SA, the search starts with a large temperature, which allows a high probability of accepting a worse solution. Then, the temperature continues to decline by a fixed value (Equation 9) as the number of iterations increases until a stopping criterion is met. In other words, as the search continues, the probability of replacing a new solution with an old solution decreases, which increases the chance of trapping in a local maxima. To overcome this, ASA is proposed, in which the annealing temperature is changed dynamically based and, in any direction, (e.g., increasing or decreasing). There are different methods proposed by [62] to control the temperature in ASA. In this study, the temperature is controlled by Equation 9 [63].

$$T_i = T_{min} + \beta \cdot \ln(1 + r_i) \tag{9}$$

Where, $T_{min}$ is the minimum temperature, $\beta$ is a constant that specifies how much the temperature can be increased, $r_i$ is the number of moves that result in a worse solution (e.g., bad moves). At any point of the search, the value of $r_i$ can take one of three values (Equation 10). If a newly generated solution ($S_{new}$) is worse than the current solution ($S_{cur}$), the value of $r_i$ is increased by 1. If the solution does not change, the value of $r_i$ remains unchanged, while if the new solution is better than the current solution the value of $r_i$ becomes 0. The rationale behind this temperature control method is that there is a better chance of accepting worse solutions at the beginning of the search and consequently a better chance of diversifying solutions. Therefore, there is less need for a high temperature to avoid a local maxima trap. However, as the search continues, the chance of trapping in a local-maxima is higher. Therefore, there is a need for high temperature to move the search out of a local-maxima. In addition, the probability of accepting a new solution is independent of the number of iterations.

$$r_i = \begin{cases} r_{i-1} + 1 & if\ f(S_{new}) < f(S_{curr}) \\ r_{i-1} & if\ f(S_{new}) = f(S_{curr}) \\ 0 & if\ f(S_{new}) > f(S_{curr}) \end{cases} \tag{10}$$

---

**Algorithm 2:** Adaptive Simulated Annealing

---

| | |
|---|---|
| 1 | Set the initial parameters of ASA |
| 2 | Generate an initial solution $S^*$ |
| 3 | Set $S^*$ as ($S_{cur}$) |
| 4 | **While** the stopping criterion is not met |
| 5 | Generate neighborhood solution ($S_{new}$) |
| 6 | Evaluate $f(S_{new})$ $and$ $f(S_{cur})$ |
| 7 | **if** $f(S_{new}) > f(S_{cur})$ |
| 8 | $S_{cur} = S_{new}$ |
| 9 | $r_i = 0$ |
| 10 | **elif** $f(S_{new}) < f(S_{cur})$ |
| 11 | $r_i = r_{i-1} + 1$ |
| 12 | **elif** $f(S_{new}) = f(S_{cur})$ |
| 13 | $r_i = r_{i-1}$ |
| 14 | Update $T_i$ and calculate $c$ |
| 15 | **if** $rand\ (0,1) <= c$ |
| 16 | $S_{cur} = S_{new}$ |
| 17 | **else** Keep $S_{cur}$ and go to line 4 |
| 18 | End |

### 3.4.3 *Integrating SA and ASA with machine learning*

Figure 4 shows the steps of using ASA for optimizing CaB. The same steps are applied for optimizing XGB. It starts by setting the parameters of ASA such as minimum temperature and number of iterations. Then, an initial solution ($S^*$) is generated, which represents the values of CaB parameters that will be optimized including learning rate, bagging temperature, tree depth, and L2 regularization (step 1). The initial values are generated using a uniform distribution with small values (e.g., 0 and 2) to avoid overfitting. Then, the initial solution ($S^*$) is set to be the current solution ($S_{cur}$) (Step 2). In step 3, a neighborhood solution is generated from the current one and called $S_{new}$. Generating a neighborhood solution is done based on the normal distribution. Each parameter value is increased or decreased by a random number generated from a normal distribution. If the parameter has a possible value greater than one (e.g., Tree depth in CatBoost), the mean and standard deviation of the normal distribution are 0 and 2, respectively. However, for a parameter that has a value between 0 and 1 such as the learning rate, the mean and standard deviation of the normal distribution are equal to 0 and 1, respectively. Therefore, when the number that is obtained from the normal distribution is negative, the current value of a parameter decreases, otherwise, it increases. This guarantees the diversification of the neighborhood solution and avoids continuously increasing or decreasing the value of each parameter.

In steps 4 and 5, the new and current solutions are examined if they violate the constraint of ASA-CaB parameters. Such violation is a learning rate larger than 1. If a parameter gets out of its limits, it is fixed by bringing it to the designated limits. For example, if a learning rate is greater than 1, its value becomes 1, but if it is less than 0, its value becomes 0.001. In steps 6, 7, and 8, the data is split into training and testing sets, and the ASA-CaB model is trained and tested based on the current $S_{curr}$ and newly generated solutions $S_{new}$. In steps 9, 10, and 11, the current solution is updated in two ways. Firstly, if the accuracy of the new solution $S_{new}$ is better than the current solution $S_{curr}$, $S_{new}$ becomes the current solution ($S_{curr} = S_{new}$) (Step 9 and 11). Secondly, a random number is generated from a uniform distribution between (0,1) and compared with a number ($c$), which is generated from Equation 7. If $c$ is greater than the random number, $S_{cur}$ is replaced by $S_{new}$. Otherwise, $S_{cur}$ remains the same and does not change. In step 12, the annealing temperature is updated according to Equation 8. This process continues until a stopping criterion is met, which is the maximum number of iterations in this work (Step 13). After trying a different number of iterations, we decide to set the number of iterations to be 500. The steps of optimizing XGB using ASA are the same as the steps for ASA-Cab. Also, the steps of using traditional SA for optimizing XGB or CaB are similar to ASA except that the annealing temperature is updated based on Equation 9, in which the initial temperature is multiplied by a fixed value $\alpha$ called cooling rate. The summary of the steps of optimizing XGB and CaB using SA or ASA is presented in Algorithm 3.

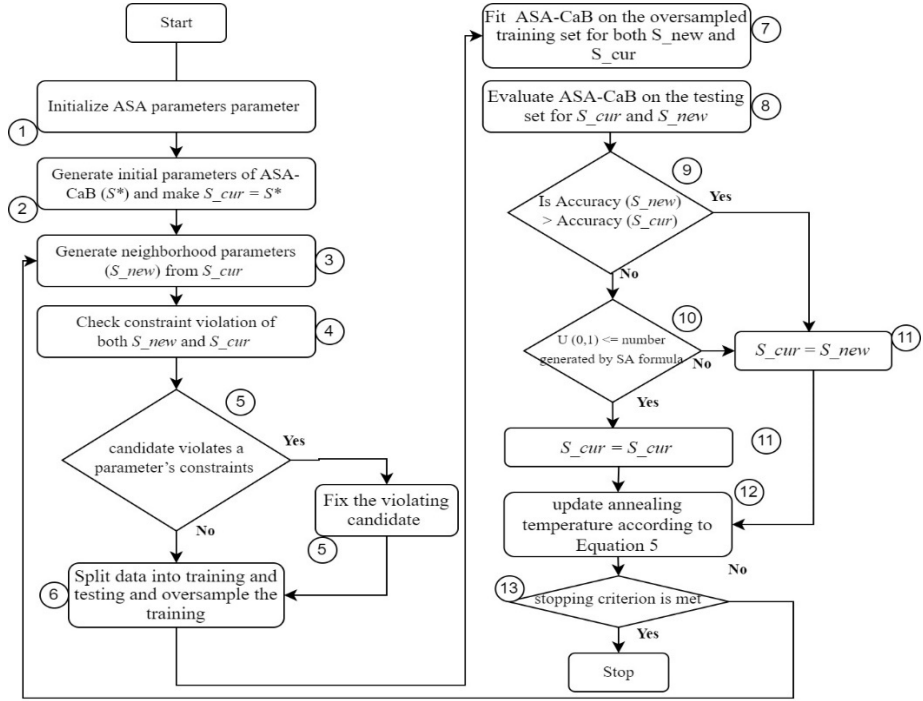

**Figure 4:** Flow chart for using ASA to optimize CaB.

| **Algorithm 3:** SA and ASA integration with XGB or CaB |
|---|
| 1.    Initialize the parameter values for machine learning algorithm (e.g., XGB or CaB) $S_{cur}$ |
| 2.    **While** a stopping criterion is not met |
| 3.       Generate a neighborhood solution from the initial solution $S_{new}$ |
| 4.       **If** $S_{cur}, S_{new}$ violates a parameter's constraints |
| 5.         Fix the violation |
| 6.       Fit the oversampled training set for both $S_{cur}, S_{new}$ |
| 7.       Predict the testing set and get accuracy for $S_{cur}, S_{new}$ |
| 8.       Update $S_{cur}$ according to algorithm 1 or 2 |
| 9.       Update the annealing temperature according to algorithm 1 or 2 |
| 10.   End |

## 3.5 Performance measures

In this work, four performance measures are utilized to evaluate the proposed models: accuracy, precision, recall, and f1 score. Equations 11 – 14 are used to calculate accuracy, precision, recall, and f1 for any class, where $n$ $(i = 1, 2, \dots, n)$ represents the number of classes, knowing that true positive (TP) for a class is the sum of all instances in which all labels are classified correctly, while true negative (TN) for a particular class is the sum of all instances where a class is classified with labels other than the target label. The TN in a confusion matrix is the sum of all the rows and columns of all the classes except the row and column of the target class. False-positive (FP) for a class is the sum of all instances located in the column of the target class in a confusion matrix except the TP. False-negative (FN) for a class is the sum of all instances located in the row of the target class in a confusion matrix except the TP.

$$Average\ Accuracy = \frac{\sum_{i=1}^{n} \frac{TP_i + TN_i}{TP_i + FN_i + FP_i + FN_i}}{n} \tag{11}$$

$$Average\ precision = \frac{\sum_{i=1}^{n} \frac{TP_i}{TP_i + FP_i}}{n} \tag{12}$$

$$Average\ recall\ = \frac{\sum_{i=1}^{n} \frac{TP_i}{TP_i + FN_i}}{n} \tag{13}$$

$$Average\ f1 = 2 * \frac{precision * recall}{precision * recall} \tag{14}$$

## 4    Experimental Results

In this section, the experimental settings for the proposed models are presented. The feature selection, as well as the prediction results, are presented in this section.

### 4.1 Experimental settings

Table 4 presents the hyperparameters that are optimized in this study for CaB and XGB along with their possible and experimental ranges. The initial values of the parameters are generated using uniform distribution and then updated using the normal distribution with mean and standard deviation 0 and 1, respectively. Five parameters are considered for CaB, which are learning rate, random strength, bagging temperature, tree depth, and L2 regularization. The default values are used for the other parameters except for the number of trees (i.e., number of iterations). The early stopping option is used to decide the number of trees. In XGB, eight parameters are optimized, in which four of them are integer and four are float. The parameters of SA-CaB and SA-XGB are optimized using SA, while the parameters of ASA-CaB and ASA-XGB are optimized using ASA. Table 5 shows the parameters of SA and ASA.

**Table 3:** Parameter ranges for CaB and XGB algorithms.

| Algorithm | Parameter | Possible range | Experimental range | Type |
|-----------|-----------|----------------|--------------------|----|
| CaB | Learning rate | (0,1] | (0,1] | Float |
| | Random strength | [1, ∞) | (0,1] | Float |
| | Bagging temperature | [0, ∞] | [0, 1] | Float |
| | Tree depth | [1,16] | [1,16] | Integer |
| | L2 regularization | [0, ∞) | [0, 14) | Float |
| XGB | Number of estimators | [1, ∞) | [1, 50] | Integer |

| | Maximum depth | $[1, \infty)$ | $[1, 50]$ | Integer |
|---|---|---|---|---|
| | Maximum Delta | $[1, \infty)$ | $[1, 50]$ | Integer |
| | Number of parallel trees | $[1, \infty)$ | $[1, 50]$ | Integer |
| | Learning rate | $(0,1]$ | $(0,1]$ | Float |
| | L1 regularization | $(0,1]$ | $(0,1]$ | Float |
| | L2 regularization | $(0,1]$ | $(0,1]$ | Float |
| | Gamma | $(0, \infty]$ | $(0,50]$ | Float |

**Table 4:** SA and ASA parameters.

| Metaheuristic | Parameter | Value |
|---|---|---|
| SA | # of iterations | 500 |
| | Initial temperature | 1000 |
| | temperature decrease ($\alpha$) | 0.8 |
| ASA | $T_{min}$ | 2 |
| | Rate of temperature increase *($\beta$)* | 2 |
| SA and ASA | Number of moves | 8 |
| | Initial solution generation | Uniform distribution |
| | Neighborhood search | Normal distribution |

**4.2 Feature selection results**

Table 7 shows the feature selection results for the seven feature selection methods used in this study. Each column represents the features that are selected by each method. The last column shows how many times a feature is selected by the seven methods. For example, the patient age feature is selected seven times by the seven methods, while patient sex is only selected once. The last row in Table 6 shows how many features are selected by each feature selection method. For example, from the 17 predictors, ten features are selected by the Lasso_SFM method, while five features are selected by DT_SFM. Table 7 gives an overview of the importance of the features. When a feature is selected not selected by any methods, it shows that the feature is not important with respect to ESI. On the other hand, when a feature is selected by all or many selection methods, it means that that feature is important with respect to ESI.

**Table 5:** Feature selection results.

| Feature | Lasso SFM | DT SFM | RF SFM | Chi SKB | DT RFE | RF RFE | Lasso RFE | Total |
|---|---|---|---|---|---|---|---|---|
| Zipcode | √ | √ | √ | √ | √ | √ | √ | 7 |
| Ed department location ID | √ | √ | √ | √ | √ | √ | √ | 7 |
| Age years | √ | √ | √ | √ | √ | √ | √ | 7 |
| Respiratory rate | √ | √ | √ | | √ | √ | √ | 6 |
| Diastolic Blood Pressure | √ | | | √ | √ | √ | √ | 5 |
| Chief Complaint | | √ | √ | √ | √ | √ | | 5 |
| Temperature In Fahrenheit | √ | | | | √ | √ | √ | 4 |
| O2 Saturation | √ | | | | √ | √ | √ | 4 |
| Systolic Blood Pressure | √ | | | √ | | | √ | 3 |
| ED Arrival Time hr | | | | √ | √ | √ | | 3 |
| BMI | | | √ | | √ | √ | | 3 |
| Pulse Rate | √ | | | | | | √ | 2 |
| Patient Ethnicity | √ | | | | | | √ | 2 |
| day of week | | | | √ | | | | 1 |
| Patient Smoking Status | | | | √ | | | | 1 |
| Patient Sex | | | | √ | | | | 1 |
| Month of year | | | | | | | | 0 |
| Total | 10 | 5 | 6 | 10 | 10 | 10 | 10 | |

## 4.3 Optimization Results

During the optimization stage, the parameters of SA-XGB, SA-CaB, are optimized using SA, while the parameters of ASA-XGB, ASA-CaB are optimized using ASA. In GS-XGB and GS-CaB, the parameters are fine-tuned using GS. The proposed algorithms (e.g., SA-XGB, SA-CaB, ASA-XGB, ASA-CaB, GS-XGB, and GS-CaB) are trained and tested using each data group that resulted from the feature selection method. Figures 5 – 8 show the convergence of the SA-XGB, SA-CaB, ASA-XGB, ASA-CaB, respectively. All the convergence graphs are based on accuracy, as it is considered as the base performance measure to decide the best model. All the models converged after about 300 iterations, knowing that all the models are run for 500 iterations. The accuracies of the converged models ranged between 65% to 83%. Tables 7 – 10 present the optimal parameters for SA-XGB, SA-CaB, ASA-XGB, ASA-CaB for all data groups, respectively. The best model among all the 48 proposed models is ASA-CaB and it is trained and tested by the data group selected by RF_RFE. For SA-XGB, the model with the highest accuracy (76%) is the model that is trained and tested by all features (X_all) (See Table 8). For SA-CaB (Table 9), the best model, which has an accuracy of 80.3%, is based on the Chi_SKB data group. For ASA-XGB (Table 10), the data group DT_RFE resulted in the best accuracy (80.3%).

**Table 6:** SA-XGB optimal parameters.

| Parameter | Lasso SFM | DT SFM | RF SFM | Chi SKB | DT RFE | RF RFE | Lasso RFE | X all |
|---|---|---|---|---|---|---|---|---|
| Number of estimators | 1.0 | 5.0 | 1.0 | 2.0 | 1.0 | 1.0 | 3.0 | 1.0 |
| Maximum depth | 1.0 | 8.0 | 1.0 | 4.0 | 5.0 | 9.0 | 1.0 | 1.0 |
| Maximum Delta | 1.0 | 9.0 | 7.0 | 1.0 | 3.0 | 3.0 | 7.0 | 3.0 |
| Number of parallel trees | 4.0 | 6.0 | 4.0 | 1.0 | 9.0 | 8.0 | 5.0 | 1.0 |
| Learning rate | 0.7 | 0.4 | 0.3 | 0.0 | 0.0 | 0.4 | 0.7 | 0.9 |
| L1 regularization (reg_alpha) | 0.3 | 0.6 | 0.3 | 0.8 | 0.8 | 0.2 | 0.6 | 0.8 |
| L2 regularization (reg_lambda) | 0.6 | 0.2 | 0.6 | 0.8 | 0.2 | 0.7 | 0.7 | 0.3 |
| Gamma | 1.0 | 44.0 | 27.1 | 20.9 | 46.0 | 13.3 | 46.7 | 2.0 |
| Accuracy | 65.3% | 70.3% | 74.2% | 75.3% | 74.0% | 73.9% | 67.3% | 76.6% |

**Table 7:** SA-CaB optimal parameters.

| Parameter | Lasso SFM | DT SFM | RF SFM | Chi SKB | DT RFE | RF RFE | Lasso RFE | X all |
|---|---|---|---|---|---|---|---|---|
| Learning rate | 0.3 | 0.8 | 0.5 | 0.6 | 0.8 | 0.4 | 0.2 | 0.4 |
| Random strength | 1.5 | 13.3 | 8.9 | 3.0 | 6.4 | 4.9 | 10.9 | 10.0 |
| Bagging temperature | 14.0 | 7.9 | 4.7 | 4.9 | 9.9 | 11.1 | 7.2 | 2.9 |
| Tree depth | 11.9 | 11.6 | 6.7 | 7.0 | 3.9 | 10.3 | 3.9 | 11.6 |
| L2 regularization | 1.0 | 1.0 | 4.0 | 1.0 | 1.0 | 1.0 | 5.0 | 1.0 |
| Accuracy | 71.7% | 78.4% | 79.1% | 80.3% | 79.1% | 77.9% | 71.5% | 79.5% |

**Table 8:** ASA-XGB optimal parameters.

| Parameter | Lasso SFM | DT SFM | RF SFM | Chi SKB | DT RFE | RF RFE | Lasso RFE | X all |
|---|---|---|---|---|---|---|---|---|
| Number of estimators | 1.0 | 1.0 | 1.0 | 6.0 | 1.0 | 1.0 | 4.0 | 7.0 |
| Maximum depth | 8.0 | 1.0 | 3.0 | 5.0 | 1.0 | 1.0 | 1.0 | 1.0 |
| Maximum Delta | 3.0 | 5.0 | 4.0 | 5.0 | 1.0 | 12.0 | 3.0 | 25.0 |
| Number of parallel trees | 2.0 | 1.0 | 1.0 | 1.0 | 5.0 | 12.0 | 2.0 | 3.0 |
| Learning rate | 0.4 | 0.2 | 0.9 | 0.2 | 0.7 | 0.0 | 0.1 | 0.9 |
| L1 regularization (reg_alpha) | 0.7 | 0.3 | 0.2 | 0.6 | 0.7 | 0.2 | 0.7 | 0.9 |
| L2 regularization (reg_lambda) | 0.0 | 1.0 | 0.4 | 0.8 | 0.8 | 0.0 | 0.3 | 0.0 |
| Gamma | 25.7 | 5.1 | 26.4 | 44.1 | 22.9 | 41.4 | 27.0 | 43.1 |
| Accuracy | 68.2% | 72.9% | 71.8% | 76.3% | 79.9% | 74.6% | 69.2% | 73.9% |

**Table 9:** ASA-CaB optimal parameters.

| Parameter | Lasso SFM | DT SFM | RF SFM | Chi SKB | DT RFE | RF RFE | Lasso RFE | X all |
|---|---|---|---|---|---|---|---|---|
| Learning rate | 0.09 | 0.17 | 0.25 | 0.99 | 0.66 | 0.05 | 0.12 | 1.00 |
| Random strength | 4.14 | 2.90 | 6.84 | 2.35 | 8.91 | 9.25 | 14.00 | 12.74 |
| Bagging temperature | 7.91 | 4.26 | 9.60 | 2.91 | 6.57 | 13.53 | 6.69 | 12.52 |
| Tree depth | 5.15 | 5.80 | 2.99 | 3.89 | 13.38 | 6.60 | 11.51 | 2.92 |
| L2 regularization | 3.00 | 2.00 | 1.00 | 3.00 | 1.00 | 2.00 | 1.00 | 1.00 |
| Accuracy | 67.4% | 80.3% | 78.1% | 83.1% | 81.9% | 83.3% (best) | 75.4% | 82.8% |

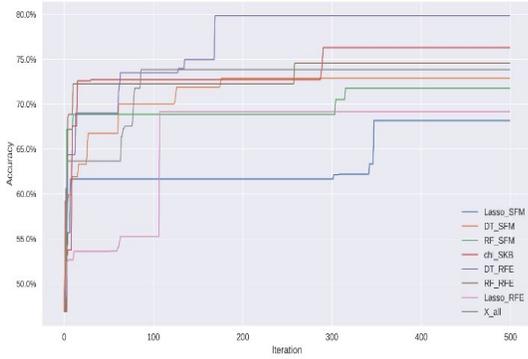

**Figure 5:** SA-XGB convergence for all groups.

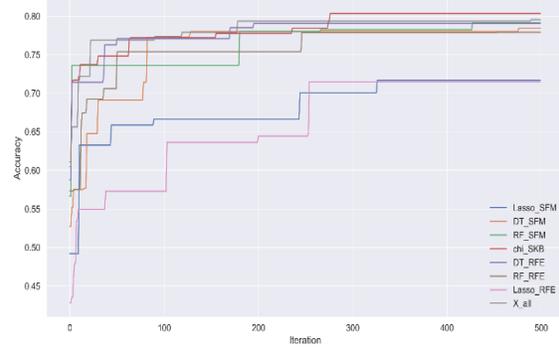

**Figure 6:** SA-CaB convergence for all groups.

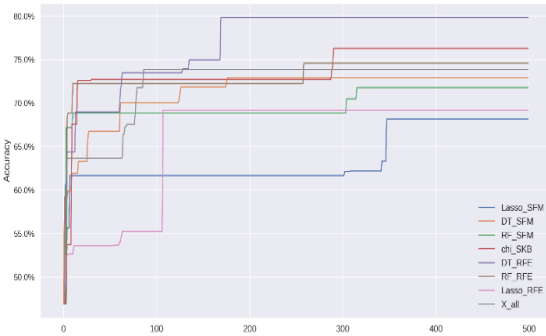

**Figure 7:** ASA-XGB convergence for all groups.

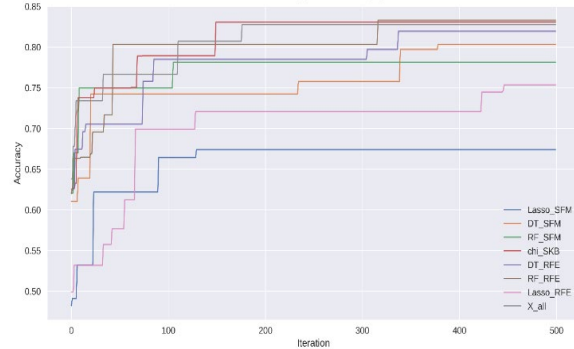

**Figure 8:** ASA-CaB convergence for all groups.

## 4.4 Prediction results

The prediction results of the testing stage for all the proposed models are presented in this section. The accuracy, precision, recall, and f1 scores for all the optimal models are shown in Figures 9 – 12, respectively. In each graph, every line represents the performance of one of the proposed algorithms including SA-XGB, ASA-XGB, SA-CaB, ASA-CaB, GS-XGB, and GS-ASA. The x-axis represents the data groups obtained from the feature selection phase, while the y-axes represent a models' performance. For the accuracies of all models (Figure 9), the ASA-CaB model that is based on the RF_RFE data group resulted in the highest accuracy (83.3%), while the GS-XGB model that is based on Lasso_RFE data group resulted in the lowest accuracy (53.3%). In terms of the optimization approach, in most cases, ASA performed better than SA and GS regardless of the data group and prediction algorithm. For example, the ASA-XGB model for Lasso_SFM, DT_SFM, and RF_SFM data groups performed better than the SA-XGB for the same data groups. The same applies to most of the data groups with respect to SA-CaB and ASA-CaB. The reason is that ASA is more likely to avoid trapping in a local maxima as the number of iterations increases because the annealing temperature is independent of the iteration number. In terms of prediction algorithm, CaB performed better than XGB in most cases regardless of the data group. For feature selection, the data group selected by RF_RFE resulted in the best, which is highlighted in Table 7. For GS, it performed worse than the SA and ASA when optimizing the two prediction algorithms (XGB and CaB). For precision, recall, and f1 scores (Figures 10 – 12), ASA-CaB performed the best among most of the data groups. The best model, which is the model that is trained and tested using ASA-CaB based on the RF_RFE data group, resulted in the highest precision, recall, and f1 scores.

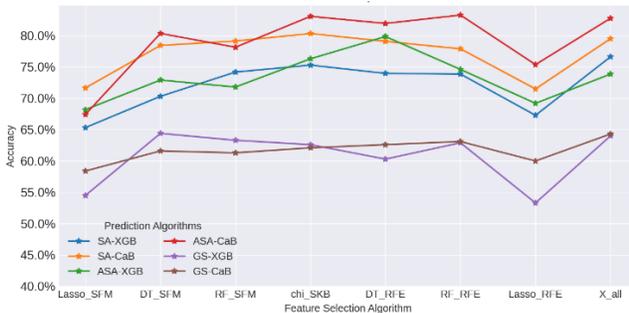

**Figure 9:** Accuracy for all prediction models.

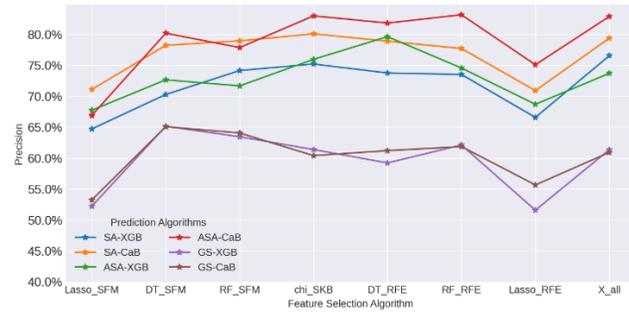

**Figure 10:** Precision for prediction models.

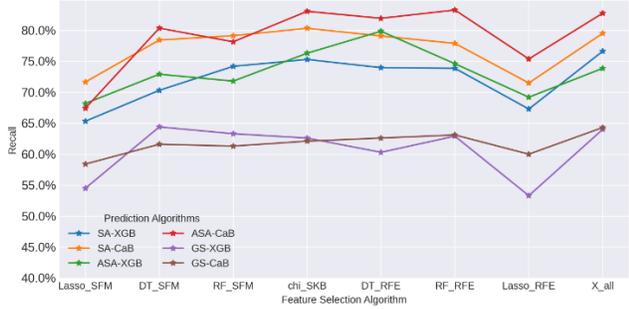

**Figure 11:** Recall for prediction models.

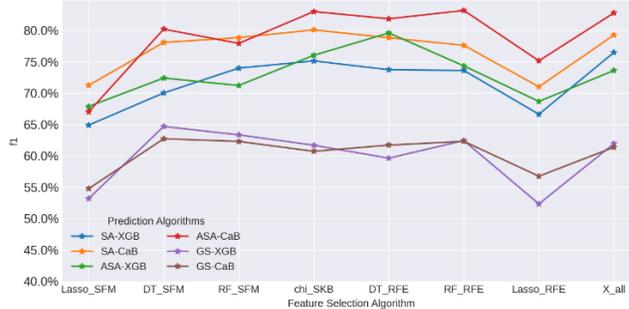

**Figure 12:** f1 score for prediction models.

### 4.5 Interpreting the predictors of the best model

The Shapley additive explanations (SHAP) value is used to interpret the effects of the features of the best model on the ESI levels. SHAP is a tool that is based on game theory to explain the impact of a feature's value on a response variable being either positive or negative [64]. Figures 13 – 15 show the SHAP graph for the best model for the ESI classes 2, 3, and 4, respectively. The X-axis represents the SHAP value, while the y-axis (left-hand side) represents the feature name. The higher the SHAP value, the higher the value of a feature.

The bar on the right-hand side represents the value of a feature. When a feature value is blue, it means a small value for the feature, and pink indicates a high value of the feature.

Figure 13 shows the effects of the model features for class 2 (most urgent). The location of ED has the highest impact, and diastolic blood pressure has the lowest effect. Most of the features negatively affect the ESI score 2. This includes ED Location ID, patient age, zip code, BMI, arrival hour, O2 saturation, and respiratory rate. As the value of these features increases, their SHAP value decreases, and consequently, the less likely a patient is assigned an ESI level 2 (most urgent). On the other hand, patient temperature positively affects ESI score 2. The higher the value of a patient temperature, the higher value of the SHAP value, and consequently the higher the chance a patient with a high temperature to be assigned an ESI score of 2 (most urgent). Figures 14 and 15 can be interpreted similarly. For example, based on Figure 15, BMI, patient age, and O2 saturation positively affect ESI level 4 (less urgent). The reason is that the higher value of these features, the higher their SHAP values. As a result, patients with high values of these features are more likely to be assigned to score 4 (less urgent). The SHAP graphs provide an interpretation CaB model.

## 5 Discussions and Implications

This work investigates an important problem with the use of machine learning in healthcare. It develops an e-triage tool based on machine learning and metaheuristic optimization. This work provides several important contributions. It proposes a framework for developing an e-triage using machine learning and metaheuristic optimization. The framework is composed of three phases, which represent the main steps for developing a machine learning model. The preprocessing steps that are needed to prepare the data for modeling are conducted in phase I. Then, the feature selection is conducted in phase II using seven methods: Lasso-SFM, DT-SFM, RF-SFM, Chi-SKB, DT-RFE, RF-RFE, Lasso-RFE. In phase III, the seven data groups that resulted from phase II along with all the features (X-all) are used to develop a model for predicting three ESI levels (2, 3, 4). The prediction algorithms are SA-XGB, ASA-XGB, SA-CaB, ASA-CaB, GS-XGB, and GS-CaB. The combination of feature selection and prediction makes 48 models. The model with the highest accuracy is selected to be the best model. During the feature selection and model development, the data is oversampled using SMOTE. The results showed that the best model is RF-RFE-ASA-XGB. This refers to RF as a feature selection algorithm and RFE as the search method that is used with RF. It also refers to the ASA metaheuristic, which is used to optimize the parameters of CaB. The accuracy of the best model is 83.2%.

This study provides theoretical and practical contributions. It shows how machine learning can be optimized using metaheuristic algorithms. Although there are existing studies that utilized metaheuristic algorithms to optimize machine learning algorithms, up to our knowledge, this study is the first study that presents a framework for optimizing machine algorithm learning (e.g., XGB and CaB) using ASA. The effectiveness and superiority of ASA over SA and GS have been shown through many experiments. In terms of practical implication, this study has accomplished the following: 1) e-triage tool is proposed based on machine learning, which can be used by nurses to assign a proper ESI level; 2) the features that affect the ESI levels are identified, which are only ten features; 3) an explanation is provided on how the ten features affect the three ESI levels; 4) Due to the high volume of patients in EDs, nurses are stressed, especially in the era of pandemic (e.g., COVID-19). The proposed tool can be used by nurses to validate their triage decision, which reduces stress. In addition, an accurate triage for ED patients helps hospitals to avoid under and over-triage and consequently improves healthcare operations and saves patients' life.

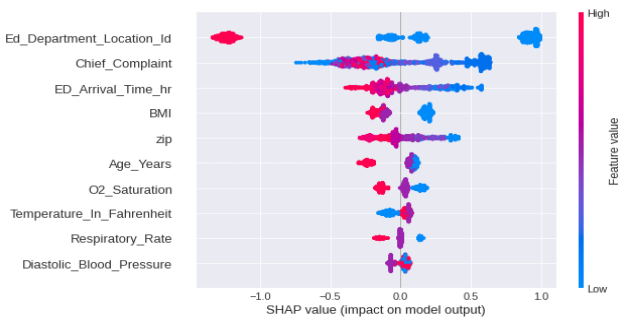

**Figure 13:** SHAP graph for ESI class 2 (most urgent).

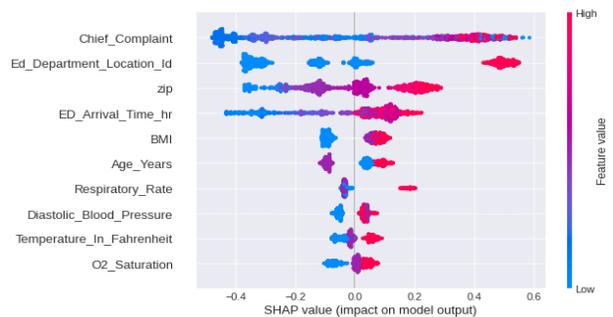

**Figure 14:** SHAP graph for ESI class 3 (urgent).

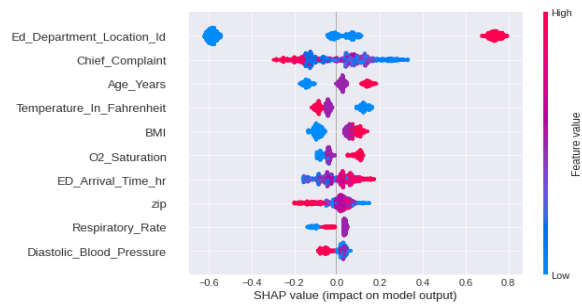

**Figure 15:** SHAP graph for ESI class 4 (less urgent).

## 6    Conclusion and Future work

This paper proposes an e-triage tool based on optimized machine learning. The tool can be used to prioritize patient care at EDs. Two main machine learning algorithms are utilized, which are XGB and CaB. The parameters of those algorithms are optimized using two optimization algorithms, which are SA and ASA. GS is also used for parameter fine-tuning to compare it with SA and ASA. Therefore, six machine learning algorithms are proposed: SA-XGB, ASA-XGB, SA-CaB, ASA-CaB, GS-XGB, and GS-CaB. Those algorithms are trained and tested using eight data groups. The findings of this work show that ASA-CaB outperforms all the other proposed algorithms. The features of the final model are interpreted using the SHAP tool, which provides not only an e-triage tool but also an explanation for the factors that affect ED patient triage. The features of the final models are ranked according to their importance with respect to the three ESI classes that are considered in this study.  Some of the key findings show that ED location is the most important feature with respect to ESI levels 2 and 4. In addition, feature interpretation reveals that some features positively affect an ESI level and negatively affect another ESI level. The positive effect means that the higher the value of a feature, the more likely a patient is assigned to a certain ESI level. The negative effect is the opposite. For example, O2 saturation negatively affects ESI level 2, and positively affects ESI level 4. This means that the higher the value of O2 saturation, the more likely a patient is assigned level 4 and less likely to be assigned to level 2. Another interesting finding is that some features are not important with respect to patient triage such as patient sex and month of a year.